\documentclass[remotesensing,article,accept,pdftex,moreauthors]{Definitions/mdpi} 
\usepackage[ruled,linesnumbered]{algorithm2e}
\newcommand{\etal}{et al.}
\firstpage{1} 
\pubvolume{17}
\issuenum{9}
\articlenumber{1556}
\pubyear{2025}
\copyrightyear{2025}
\externaleditor{Ming-Der Yang} % More than 1 editor, please add `` and '' before the last editor name
\datereceived{6 February 2025} 
\daterevised{21 April 2025} % Comment out if no revised date
\dateaccepted{25 April 2025} 
\datepublished{27 April 2025} 
\hreflink{https://doi.org/10.3390/\\rs17091556} % If needed use \linebreak
\Title{AD-Det: Boosting Object Detection in UAV Images with Focused Small Objects and Balanced Tail Classes}

\TitleCitation{AD-Det: Boosting Object Detection in UAV Images with Focused Small Objects and Balanced Tail Classes}

\Author{{Zhenteng Li} %MDPI: Please carefully check the accuracy of names and affiliations.
 $^{1}$, Sheng Lian $^{1,2,3,}$*\orcidA{}, Dengfeng Pan $^{1}$, Youlin Wang $^{1}$ and Wei Liu $^{4}$\orcidB{}}

\AuthorNames{Zhenteng Li, Sheng Lian, Dengfeng Pan, Youlin Wang and Wei Liu}

\isAPAStyle{%
       \AuthorCitation{Lastname, F., Lastname, F., \& Lastname, F.}
         }{%
        \isChicagoStyle{%
        \AuthorCitation{Lastname, Firstname, Firstname Lastname, and Firstname Lastname.}
        }{
        \AuthorCitation{Li, Z.; Lian, S.; Pan, D.; Wang, Y.; Liu, W.}
        }
}

\address{%
$^{1}$ \quad College of Computer and Data Science, Fuzhou University, Fuzhou 350108, China\\
$^{2}$ \quad Engineering Research Center of Big Data Intelligence, Ministry of Education, Fuzhou 350002, China
\\
$^{3}$ \quad Key Laboratory of Intelligent Metro of Universities in Fujian, Fuzhou 350108, China  \\
$^{4}$ \quad  School of Software and Information Engineering, East China Jiaotong University, Nanchang 330013, China; {weiliu@ecjtu.edu.cn}

}
\corres{Correspondence: shenglian@fzu.edu.cn}

\abstract{Object detection in unmanned aerial vehicle (UAV) images poses significant challenges due to complex scale variations and class imbalance among objects. 
Existing methods often address these challenges separately, overlooking the intricate nature of UAV images and the potential synergy between them. In response, this paper proposes AD-Det, a novel framework employing a coherent coarse-to-fine strategy that seamlessly integrates two pivotal components: adaptive small object enhancement (ASOE) and dynamic class-balanced copy--paste (DCC).
ASOE utilizes a high-resolution feature map to identify and cluster regions containing small objects. These regions are subsequently enlarged and processed by a fine-grained detector. On the other hand, DCC conducts object-level resampling by dynamically pasting tail classes around the cluster centers obtained by ASOE, maintaining a dynamic memory bank for each tail class. This approach enables AD-Det to not only extract regions with small objects for precise detection but also dynamically perform reasonable resampling for tail-class objects. Consequently, AD-Det enhances the overall detection performance by addressing the challenges of scale variations and class imbalance in UAV images through a synergistic and adaptive framework.
We extensively evaluate our approach on two public datasets, i.e., VisDrone and UAVDT, and demonstrate that AD-Det significantly outperforms existing competitive alternatives. 
Notably, AD-Det achieves a $37.5\%$ average precision (AP) on the VisDrone dataset, {surpassing its counterparts by at least $3.1\%$.} 
}


\begin{document}

\section{Introduction}
Unmanned aerial vehicles (UAVs) have experienced widespread adoption in recent years, driven by their cost-effectiveness and adaptability. Outfitted with cameras and various sensors, UAVs find application in diverse scenarios such as surveillance, rescue operations, agriculture, express delivery, and more. The effectiveness of these applications relies heavily on the seamless and robust recognition of objects within the UAV's field of view, as noted in recent comprehensive reviews ~\citep{osco2021review,zuo2022unmanned,kurunathan2023machine}.

Solutions including R-CNN series~\citep{girshick2014rich,ren2015faster}, YOLO series~\citep{redmon2016you,wang2023yolov7} and, {most recently, a query-based series}~\citep{gao2022adamixer,li2022r,jia2023detrs} have shown leading performance on public datasets such as MS COCO~\citep{lin2014microsoft}.
Despite these accomplishments, when applied to UAV images, particularly for datasets like VisDrone \citep{zhu2021detection}, {generic detectors' performance falls short of satisfactory in terms of accuracy and efficiency.} Challenges arise due to hardware limitations, complexities in the imaging environment, and flight trajectory intricacies, leading to the following issues.

\begin{figure}[H]
    \includegraphics[width=0.99\textwidth]{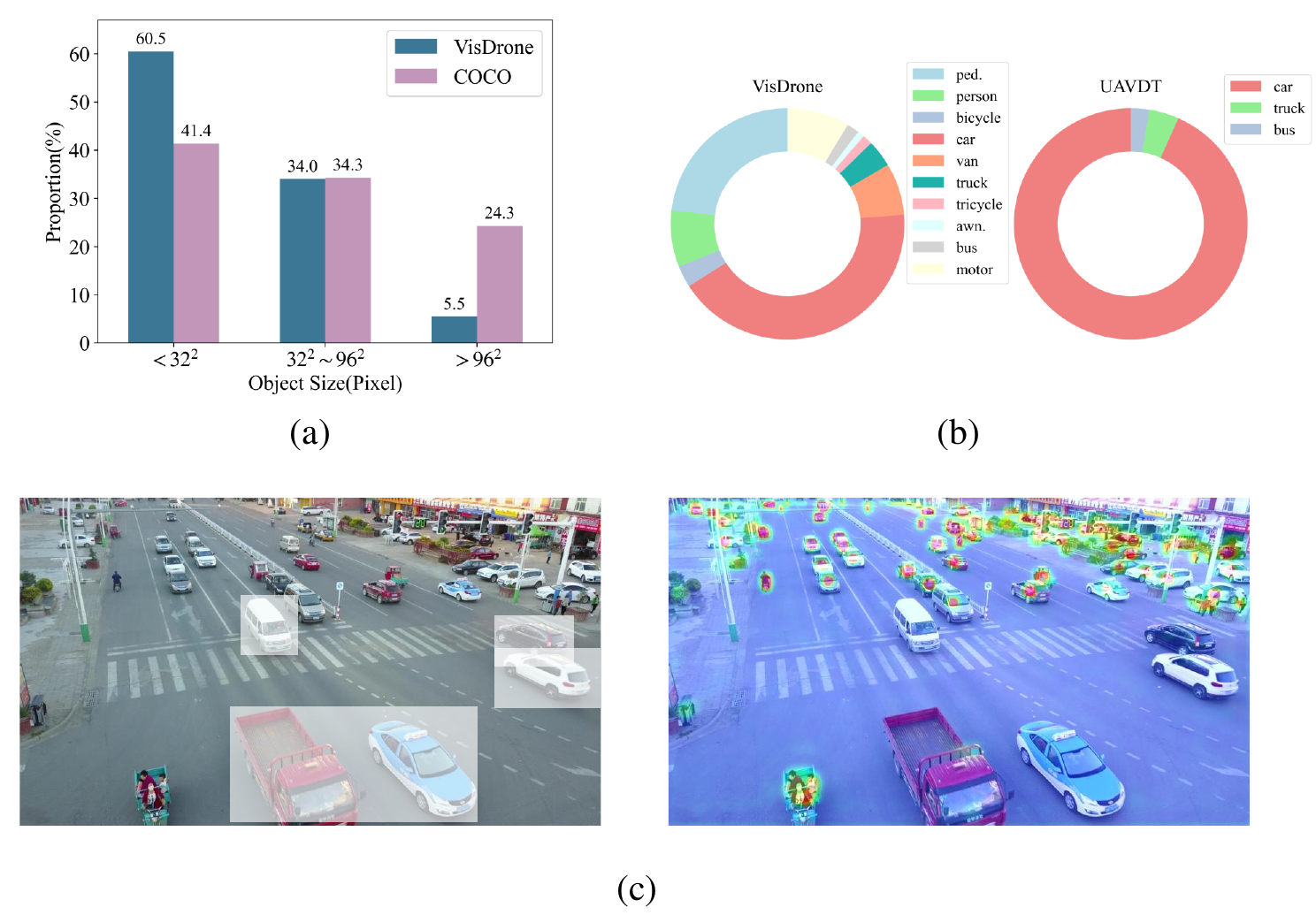}
    \caption{(\textbf{a}) {Comparison} %MDPI: 1.Please confirm whether an explanation of colors needs to be added to the figure caption. 2. Please remove the non-English term from the figure or add a definition for it.
 of scale distribution between VisDrone and COCO. (\textbf{b}) Class distribution of VisDrone and UAVDT. (\textbf{c}) Visualization of high-resolution feature map ($P_3$) in GFL~\citep{li2020generalized} for VisDrone. It can be seen that the high-resolution feature map mainly focuses on small objects. 
    The masked areas on the left indicate regions that contain large objects, which are easier to handle and can be ignored in the fine-grained detection stage.}
    \label{fig:challenges}
\end{figure}

\textbf{{Scale variations:} %MDPI: 1. Please confirm if the bold formatting is necessary; if not, please remove it. The following highlights are the same. 2. Please confirm if keep noindent format. The following highlights are the same. 
} 
{Owing to variations in flight altitudes and complex object distribution in UAV images, object sizes differ by distance from camera exposure, exhibiting a broad range of scales, with a predominance of small-scale objects.}
As illustrated in Figure~\ref{fig:challenges}a, 
within the VisDrone dataset, the proportion of small objects (object size $<32^2$ pixels) reaches $60.5\%$, which is $19.1\%$ higher than that in the COCO dataset. 
On the other hand, the percentage of large objects (object size $>96^2$ pixels) accounts for only $5.5\%$.
The limited feature representation of small objects poses a considerable challenge, which leads to decreased accuracy and reliability in detecting smaller objects, thus significantly affecting the overall detection performance.

\textbf{{Class imbalance:} 
} 
% Since the majority of UAV images are captured in urban scenarios such as VisDrone, specific categories, such as \textit{car} and \textit{people}, dominate and account for more than $70\%$ of all objects.
% Conversely, the other categories, such as \textit{bicycle} and \textit{van}, constitute a mere $1\%$ to $7\%$ 
%  (Figure~\ref{fig:challenges}(b)).
% This severe class imbalance leads the detection model to inherently favor learning from the majority classes, resulting in subpar detection performance for the classes with smaller proportions.
% {UAV images collected in urban environments exhibit a long-tailed class imbalance characteristic: a small subset of "head" classes (e.g., car and people) dominates the dataset, collectively representing over 70\% of object instances, while numerous "tail" classes (e.g., bicycle and van) are severely underrepresented, constituting only 1\% to 7\% of instances (Figure 1(b)). This skewed distribution creates a model bias towards head classes during training—detectors prioritize optimizing features for high-frequency categories at the expense of tail classes. Consequently, tail classes suffer from poor localization accuracy and high false-negative rates due to insufficient discriminative feature learning.}
{UAV images collected in urban environments exhibit a long-tailed class imbalance characteristic; a small subset of head classes (e.g., car and people) dominates the dataset, collectively representing over 70\% of object instances, while numerous tail classes (e.g., bicycle and van) are severely underrepresented, constituting only 1\% to 7\% of instances (Figure~\ref{fig:challenges}b). This skewed distribution creates a model bias towards head classes during training—detectors prioritize optimizing features for high-frequency categories at the expense of tail classes. Consequently, tail classes suffer from poor localization accuracy and high false-negative rates due to insufficient discriminative feature learning.}

To address these challenges, researchers have made many attempts. 
For scale variations, {ref.} %MDPI: Newly added  information. Please confirm. The following highlights are the same
 \cite{gao2023global, chen2023high,liu2024multitask,liu2023attention} focuses on multi-scale feature fusion, and {ref.} \cite{lin2017focal,biswas2022small, biswas2024unsupervised} emphasizes hard sample mining. 
Nevertheless, due to the input scale limitation, there is still room for improvement in accuracy.
Another direct method involves cropping the image into subregions before applying detection, such as in uniform cropping. 
However, these cropping strategies cannot adaptively adjust based on the semantic information in UAV images, potentially including extensive background areas.
Li \etal~\cite{li2020density} utilize a density map to guide the cropping of images, enhancing the semantics of the resulting subregions. 
However, this approach involves a density map generation network, which increases model complexity and requires ground truth density map generation.
Compared to the scale variations, the challenge of class imbalance is often neglected in UAV images. 
% {In recent years, research in the field of long-tailed detection has primarily been categorized into three types: (1) resampling and data augmentation techniques (such as Copy-Paste); (2) loss function reweighting methods (such as Focal Loss); and (3) decoupled optimization strategies (as proposed by Kang et al., involving two-stage training). However, existing approaches still face challenges in dynamically adapting to changes in the distribution of tail classes.}
{In recent years, research in the field of long-tailed detection has primarily been categorized into three types: (1) resampling and data augmentation techniques~\cite{Ghiasi_2021_CVPR}; (2) loss function reweighting methods~\cite{lin2017focal,Li_Liu_Wang_2019}; and \mbox{(3) decoupled} optimization strategies~\cite{kang2019decoupling}. However, existing approaches still face challenges in dynamically adapting to changes in the distribution of tail classes.}
Zhang \etal~\cite{zhang2019dense} employ a multi-model fusion (MMF) strategy to tackle the head and tail classes distinctly, effectively enhancing the detection performance of tail classes. 
However, this approach results in the discarding of a substantial amount of valuable data during the training of each model, which could potentially compromise the model’s \mbox{representational capacity.}

In this paper, for object detection in UAV images, we propose a novel framework called AD-Det, which adopts a coarse-to-fine strategy and mainly consists of two key components: adaptive small object enhancement (ASOE) and dynamic class-balanced copy--paste (DCC).
Specifically, ASOE employs a high-resolution feature map from the classification head to pinpoint small objects. 
% (Figure~\ref{fig:challenges}(c)). 
As shown in Figure~\ref{fig:challenges}c, most small objects in the image show significant activation in the high-resolution feature map.
These positions of interest are clustered into $\mathcal{N}$ subregions, which are then enlarged and processed by a fine-grained detector, thereby enhancing the detection performance of small objects.
DCC, on the other hand, performs object-level resampling through a dynamic copy-and-paste strategy specifically tailored for tail-class objects.
It leverages the clustering cues from ASOE to dynamically search for suitable pasting positions around the cluster center and maintain a dynamic memory bank for each tail class.
Additionally, data augmentation is conducted to avoid overfitting.
In this way, the proposed AD-Det can extract regions with small objects for fine-grained detection and dynamically perform reasonable resampling for tail-class objects, thereby improving overall detection performance.  

Compared to existing coarse-to-fine solutions, our approach is a plug-and-play, unsupervised scheme. The advantages lie in leveraging the inherent properties of the detector to focus on small object regions that truly contribute to accuracy gains, thereby enhancing the precision of small object detection without extra learnable parameters. Additionally, we propose a training-only image augmentation technique for tail classes, which helps to alleviate class imbalance issues.

We summarize four main contributions of this work as follows.
\begin{itemize} %
    \item We propose AD-Det, a novel object detection framework that addresses two key challenges in UAV image object detection---complex scale variations and class imbalance---ultimately enhancing detection performance.
    \item To enhance small object detection, we propose ASOE, which uses a high-resolution feature map to cluster regions containing small objects, followed by processing them for fine-grained detection.
    \item To enhance tail-class object detection, we propose DCC, which performs reasonable object-level resampling by dynamically pasting tail classes around clustering centers obtained by ASOE.
    % \item {Experimental results..}   % 补上实验的Lancer
    \item We conduct extensive experiments on VisDrone and UAVDT, which demonstrate that our approach significantly outperforms existing competitive alternatives.
\end{itemize}

\section{Related Works}

In object detection tasks, unlike traditional solutions that rely on handcrafted features and classifiers, DL-based solutions can automatically learn discriminative features and achieve promising results.
Here, we briefly review object detection solutions in the deep learning era, including (1) generic object detection (in Section~\ref{subsec:related-generic}) that focuses on natural images such as MS COCO~\citep{lin2014microsoft}; (2) object detection in aerial images (in Section~\ref{subsec:related-uav}) that focuses on aerial images such as VisDrone~\citep{zhu2021detection} and DOTA~\cite{ding2021dota}; and (3) long-tailed object detection (in Section~\ref{subsec:related-Long-tail}) that focuses on long-tail scenarios such as LVIS~\citep{gupta2019lvis}.

\subsection{Generic Object Detection}
\label{subsec:related-generic}

Sparked by impressive success in image classification tasks~\citep{krizhevsky2017imagenet}, DL-based methods have dominated generic object detection.
% According to the detection pipeline, existing solutions can be roughly categorized into two types:
The existing DL-based detection solutions, when analyzed according to their pipeline, can be roughly grouped into two-stage and one-stage methods.
The two-stage detectors are represented by R-CNN~\citep{girshick2014rich}, Fast R-CNN~\citep{girshick2015fast}, Faster R-CNN~\citep{ren2015faster}, and Mask R-CNN~\citep{he2017mask}.
These detectors first reduce the search space significantly by extracting candidate region proposals.
The extracted proposals are then classified into specific categories and refined to proper locations.
Such a pipeline leads to relatively higher accuracy but lower efficiency.
Different from the above methods, one-stage methods, on the other hand, directly predict objects' categories and locations without region proposals.
The representative approaches include  
the YOLO series~\citep{redmon2016you,redmon2017yolo9000,wang2023yolov7}, 
SSD~\citep{liu2016ssd}, and RetinaNet~\citep{lin2017focal}. 
Such a design leads to relatively higher efficiency but at the cost of accuracy.
Recently, studies into CornerNet~\citep{law2018cornernet} and FCOS~\citep{tian2019fcos} bypassed the anchor boxes mechanism and corresponding hyperparameter setting and presented promising alternatives for one-stage methods.
DETR~\citep{carion2020end} pioneered a fully end-to-end object detector that employed a transformer-based architecture, eliminating reliance on anchor generation and non-maximum suppression (NMS).
Although these detectors achieved impressive progress in generic object detection, their performance on UAV images is far from satisfying due to problems of scale variations and class imbalance.

\subsection{Object Detection in Aerial Images}
\label{subsec:related-uav}
Compared with generic objects that are mostly captured in ground view,
object detection in UAV images presents heightened challenges due to object scale and object/category distribution problems.

Many studies focus on \textbf{{multi-scale/tiny-scale object problems} 
} in aerial images. 
Most of the early research focuses on the migration of classical generic object detection algorithms.
% 增加了一些所有早些的引用注释了
% For example, \cite{sommer2017fast} investigated the application of Fast / Faster R-CNN for vehicle detection in aerial images.
% The region proposal processes in these two models are improved to handle tiny instances and generate better candidate regions.
% \cite{zhang2019scale} gave a similar scale adaptive solution, where the multi-layer RPNs were proposed for optimizing Faster R-CNN and generating multi-scale region proposals. 
% \cite{deng2017toward} combined vehicle attribute learning network with vehicle proposal network.
% Such design integrated vehicle type and orientation information into the detection model to improve the detection accuracy of small vehicles. 
% Recently, 
Yang \etal~\cite{yang2022querydet} proposed a novel query mechanism, which effectively and efficiently incorporates an additional high-resolution layer to promote small object detection.
% \cite{zhu2023scalekd} improved the performance of small object detection by combining a novel Scale-aware Knowledge Distillation (ScaleKD), which transferred the scale-aware knowledge from a complex teacher model to a compact student model.
In the realm of knowledge distillation, Zhu \etal ~\cite{zhu2023scalekd} costlessly enhanced the performance of lightweight models by incorporating scale-aware knowledge from more complex ones.
To detect small weak objects in UAV images, Han \etal ~\cite{han2022context} presented a context--scale-aware detector, combining the strengths of context-aware learning and multi-scale feature extraction.
% Gao \etal~\cite{gao2023global} designed a scale-aware network to enhance the contrast between the background and foreground objects and mitigate the impact of noise.
% Based on \cite{biswas2022small}, Biswas \etal ~\cite{biswas2024unsupervised} presented support-guided debiased contrastive learning to remove the bias on domain classes and improve the robust of model.
Cao \etal~\cite{cao2024visible} introduced a self-reconstructed difference map approach to enhance feature visibility for challenging tiny object detection tasks.
DQ-DETR~\citep{huang2024dq} specialized in tiny object detection by employing dynamic query selection and counting-guided feature enhancement, boosting detection performance.
SDPDet~\citep{yin2024sdpdet} employed scale-separated dynamic proposals and activation pyramids to enhance the efficiency and accuracy of object detection in UAV views.
{Zhang~\etal~\cite{zhang2023boost} enabled detectors to focus on discriminative features while reducing false-positives in cluttered backgrounds.
Zhang~\etal~\cite{zhang2024integrally} proposed IMPR-Det, which can integrally mix multi-scale pyramid representations for different components of an instance.}

Besides object scale problems, many studies focus on \textbf{{object/category distribution problems}}.
% 注释早些时候的引用
{Inspired by research into crowd counting, Li~\etal~\cite{li2020density} injected estimation of density maps into a conventional object detection framework, which was utilized to predict object distribution, reduce the influence of background, and generate more balanced candidate proposals. 
Duan~\etal~\cite{duan2021coarse} further adopted a coarse-grained density map to identify subregions more accurately.
Similar observations regarding the object distribution issues led \cite{yang2019clustered} to improved RPN in order to cluster region proposals. GLSAN~\cite{deng2021global} utilized a scale-aware {algorithm} to fuse global and local detection results. }
CZDet~\cite{meethal2023cascaded} enhanced cascade detection by innovatively incorporating high-density subregion labels into the repurposed detector.
YOLC~\cite{liu2024yolc} employed a local scale module and deformable convolutions to enhance accuracy in aerial small object detection.
% CEASC~\citep{du2023adaptive} optimized object detection on UAV images by using context-enhanced group normalization and adaptive multi-layer masking to improve accuracy and efficiency.
% Xu \etal~\cite{xu2023dynamic} proposed a dynamic coarse-to-fine learning (DCFL) approach for detecting oriented tiny objects, enhancing detection accuracy through refined label assignment and feature prior matching.
% CFINet~\citep{yuan2023small} enhanced small object detection through a two-stage coarse-to-fine proposal mechanism and feature imitation learning.
% Chen \etal~\cite{chen2023high} proposed HR-FPN, featuring high-resolution feature alignment and fusion along with a multi-scale decoupled head, to enhance small object detection from drone imagery.
{Zhang~\cite{zhang2024struct} proposed structured adversarial self-supervised pretraining to strengthen both clean accuracy and adversarial robustness.}
% Ren \etal~\cite{ren2022improved} enhanced Mask R-CNN for UAV thermal infrared object detection, integrating a more efficient backbone and prior knowledge filtering, thereby boosting processing speed and reducing storage needs.
% On the other hand, \cite{yu2021towards} investigated the long-tail category distribution issues, where the class-biased samplers and bilateral box heads are proposed to deal with tail classes and head classes in a dual-path manner.
On the other hand, Yu \etal~\cite{yu2021towards} investigated long-tail category distribution issues, where they designed dedicated samplers and detection heads to address the distinct characteristics of tail and head classes.

% {Different from xxxx}
Such excellent works have indeed facilitated aerial image object detection. 
Nevertheless, when addressing scale variations and class imbalance challenges, the majority of preceding studies treat these essential issues rigidly and separately, not only ignoring the complexity of UAV images but also neglecting their potential synergy.

\subsection{Long-Tail Object Detection}
\label{subsec:related-Long-tail}

Much like the paradigm of long-tail classification, research in the domain of long-tail object detection predominantly adheres to two distinctive methodologies: resampling and reweighting.
Commonly employed techniques in resampling involve either oversampling the minority classes or undersampling the majority classes during the training process, aimed at mitigating the issue of imbalanced class distribution.  
% Repeat factor sampling (RFS)~\citep{gupta2019lvis} oversampled the training data from tail classes while undersampled those from head classes from the image level. 
Repeat factor sampling (RFS)~\citep{gupta2019lvis} involves an image-level sampling approach following the resampling paradigm.
SimCal~\citep{wang2020devil} proposed a bi-level class balanced sampling approach to alleviate classification head bias. 
In the context of reweighting, the fundamental concept lies in the assignment of diverse weights to training samples, with a focus on enhancing the training of tail samples. 
Cui \etal~\cite{cui2019class} introduced a novel loss function that significantly improves class imbalance handling by dynamically adjusting the weights of training samples based on their \mbox{effective numbers.}

Besides the aforementioned strategies, many researchers have endeavored to handle the long-tail problem from other perspectives. 
Kang \etal~\cite{kang2019decoupling} advocated for a decoupled training paradigm that disentangles the learning process into distinct phases: representation learning and classifier training.  
% BAGS~\citep{li2020overcoming} grouped classes based on their class-wise instance numbers and subsequently applied softmax loss within each group. 
BAGS~\citep{li2020overcoming} grouped classes based on their instance frequency and subsequently applied softmax loss among each group.
ROG~\citep{zhang2023reconciling} proposed a multi-task learning approach that concurrently optimizes both object-level classification and global-level score ranking. 
Hyun \etal~\cite{hyun2022long} introduced effective class-margin loss (ECM) as a novel surrogate objective through which to optimize margin-based binary classification error on the imbalanced training set. 

Many earlier coarse-to-fine counterparts pinpoint regions of interest by relying on extra learnable modules or by post-processing sparse boxes. 
Our method, however, differs by incorporating dense indicators from a high-resolution feature map. 
This map {is} adept at preserving extensive information about small objects, facilitating adaptive region generation. 
Additionally, we enhance this framework by addressing class imbalance issues through the integration of object-level copy--paste techniques. 
This novel combination effectively tackles two distinct challenges, which are overlooked in existing solutions.

\section{Methodology}

% In this section, we elaborate on the design of AD-Det, including the overall framework of AD-Det in Sec.~\ref{subsec:frmoverview}, the ASOE module in Sec.~\ref{subsec:ASOE}, the DCC module in Sec.~\ref{subsec:dcc}, along with the objective functions in AD-Det's learning procedure in Sec.~\ref{subsec:train-infer-details}.

In this section, we elaborate on the design of AD-Det, including the overall framework of AD-Det in Section~\ref{subsec:frmoverview}, the ASOE module in Section~\ref{subsec:ASOE}, and the DCC module in Section~\ref{subsec:dcc}, 
along with the training and inference details of AD-Det in Section~\ref{subsec:train-infer-details}.
% \textcolor{red}{along with the training and inference details of AD-Det in Sec.~\ref{subsec:train-infer-details}.}

\subsection{Framework Overview}
\label{subsec:frmoverview}

\begin{figure}[H]
%    \centering    
    \includegraphics[width=.99\textwidth]{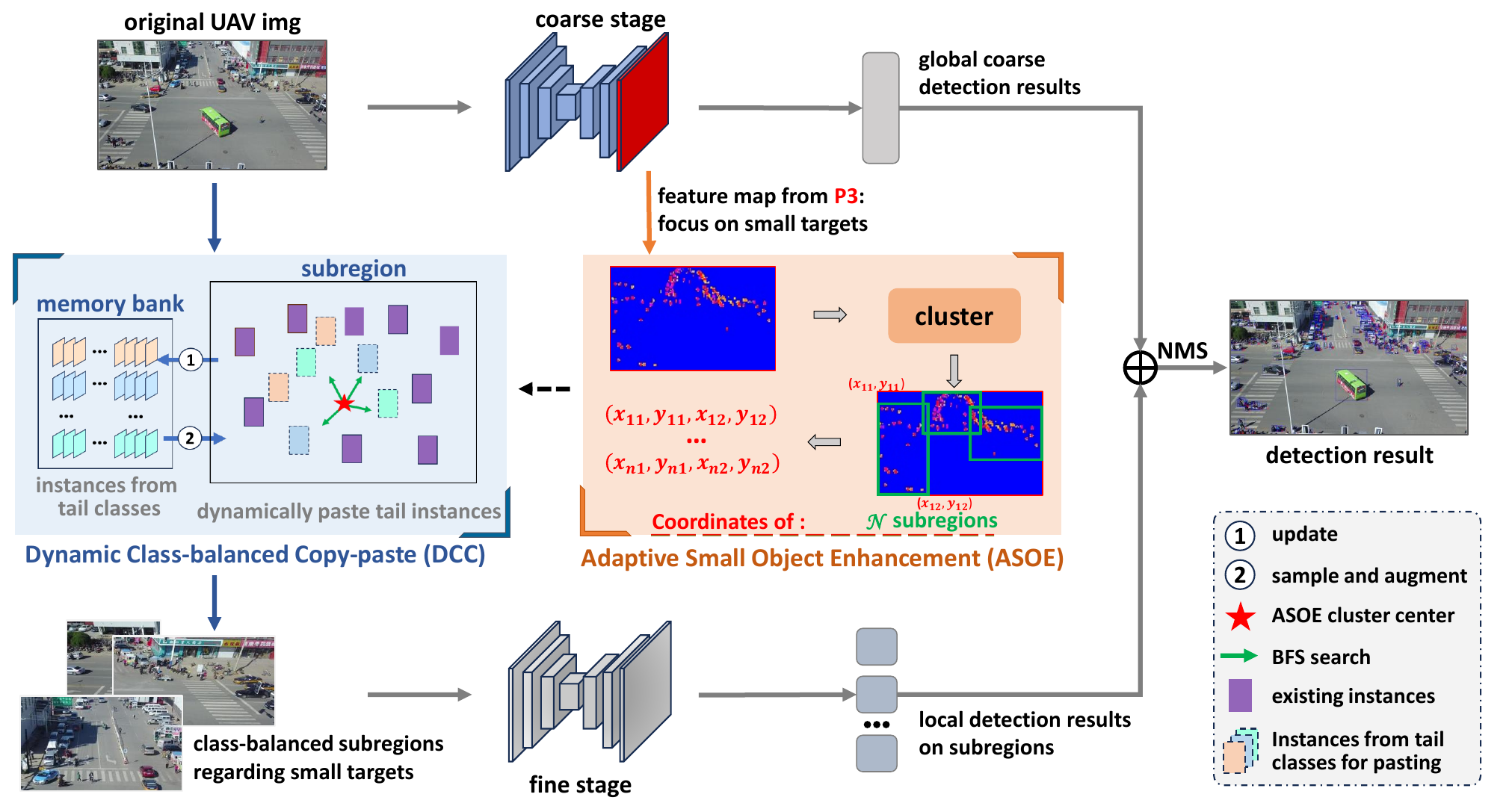}
    \caption{{The} %MDPI: We moved all figures/tables below the first mention, please confirm
 framework overview of the proposed AD-Det. 
    The network adopts a coarse-to-fine strategy and integrates two key modules, i.e., an adaptive small object enhancement (ASOE) module for excavating regions containing small objects and a dynamic class-balanced copy--paste (DCC) module for balancing class distribution. 
    The final results are obtained by fusing global results and local results using non-maximum suppression (NMS). The dashed line between ASOE and DCC indicates training-only.}
    \label{fig:framework}
\end{figure}

To tackle the key challenges of object detection in UAV images, i.e., scale variations and class imbalance, a novel approach namely AD-Det is proposed.  
Designed in a coarse-to-fine manner, AD-Det mainly consists of the following two key components.
\textbf{{Adaptive small object enhancement (ASOE)}} utilizes the coarse detector cues to roughly locate small objects, and cluster the region of interest for later fine-grained detection. 
\textbf{{Dynamic class-balanced copy--paste (DCC)}} performs reasonable object-level resampling by dynamically pasting tail classes around the cluster centers obtained by ASOE.
% combines cluster centers learned from ASOE to dynamically paste tail classes for balancing the class distribution. 
The prediction given by the coarse-to-fine strategy is fused through non-maximum suppression (NMS). 
% The overall framework of the proposed method is illustrated in 
We illustrate the overall framework of AD-Det in
Figure~\ref{fig:framework}.

\subsection{Adaptive Small Object Enhancement (ASOE)}
\label{subsec:ASOE}
\label{subsec:small-enhance}

\begin{figure}[H]
    \centering    
    \includegraphics[width=\textwidth]{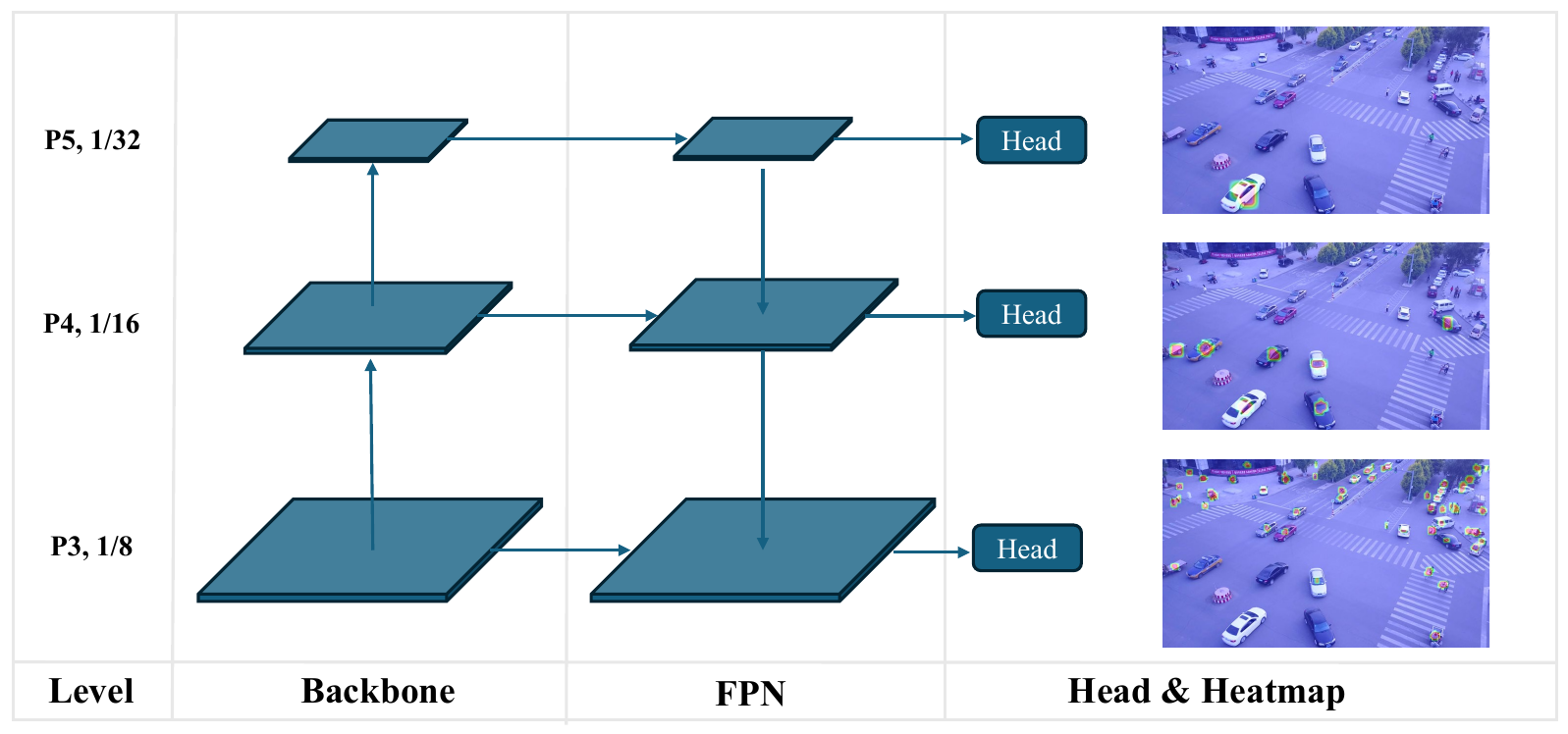}
    \caption{{Detector} %MDPI: Please confirm whether an explanation of colors needs to be added to the figure caption.
 workflow and corresponding heatmap visualizations at different network layers.}
    \label{fig:net_heatmap}
\end{figure}

In UAV images, small objects constitute a large proportion and are sparsely distributed (Figure~\ref{fig:challenges}a,c), resulting in unsatisfactory detection performance. 
Such small objects are typically detected from high-resolution low-level feature maps, for example, $P_3$ in FPN~\citep{lin2017feature}.
As shown in Figure~\ref{fig:net_heatmap}, the image is passed through the backbone to extract multi-scale features, which are subsequently fused in the FPN. 
These fused features are then passed to the detection head.  
{Due to varying convolutional strides in different layers}, the feature map sizes decrease from $P_3$ to $P_5$. 
The lower layers, having undergone fewer downsampling operations, not only retain more local information but also have a receptive field better aligned with small objects.
As observed from the heatmap comparison, small objects are primarily detected at the $P_3$ layer.
% {we discuss this in Sec.~\ref{subsub:ablation-layer}.}
Specifically, the human visual system excels at object detection by quickly scanning the entire image to acquire information on large objects and gain valuable clues in challenging areas with small objects.

Inspired by these factors, we propose adaptive small object enhancement (ASOE), which utilizes a high-resolution feature map to pinpoint regions with small, difficult-to-detect objects for fine-grained detection.
Unlike \citep{li2020density}, which relies on an additional density network to obtain the position of the foreground object, the proposed ASOE module employs a hierarchical approach, seamlessly integrating coarse-to-fine strategies to overcome the inherent challenges of identifying small objects in UAV images.

% effectively address the challenges posed by small objects in UAV images. 

\begin{figure}[H]
    \centering    
    \includegraphics[width=\textwidth]{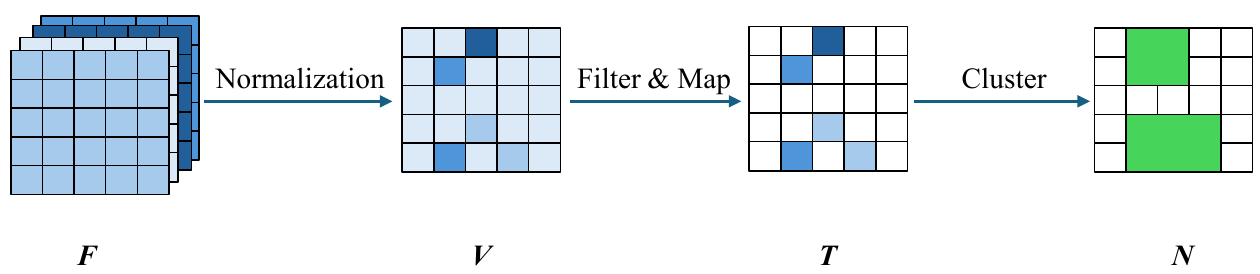}
    \caption{{The} %MDPI:  Please confirm whether an explanation of colors needs to be added to the figure caption.
 process of subregion generation in ASOE.}
    \label{fig:progress}
\end{figure}

In the first stage, a detector $Coarse(\cdot)$ captures the global context and produces preliminary detection results. 
Meanwhile, the feature map $F_{l}\in\mathbb{R}^{C\times H\times W}$ (Figure~\ref{fig:framework} coarse stage {red} layer), targeting semantic details pertinent to small objects, is extracted from the classification head enriched with high-resolution and low-level features (Figure~\ref{fig:net_heatmap}), akin to the methodology employed by RetinaNet~\citep{lin2017focal}.
Here, $l$ represents the layer index of $P_l$, while $C$, $H$, and $W$ represent the channel count, height, and width of the feature map. 
{As shown in Figure~\ref{fig:progress}, we compute a position-wise activation map $V_{l}\in\mathbb{R}^{1\times H\times W}$ by applying the Sigmoid function independently to each channel of $F_l$, followed by averaging\mbox{ across{ channels: }} %MDPI:   This manuscript has many formulas, please check if all the same content formatting is consistent (italic or not, bold or non-bold), if not, we recommend revising them to be consistent
%MDPI:  Please check to make sure all formulas are not duplicated
\begin{equation}
    V_l = \frac{1}{C}\sum_{c=1}^{C}\sigma(F_l),
\end{equation}
where $V_l(i,j)$ indicates the probability of the grid$(i, j)$ containing a small object, and $\sigma$ is the Sigmoid function. 
To suppress background noise, we retain only positions where $V_l(i,j)$ exceeds a threshold $\gamma$ = 0.5, and by multiplying with the downsampling factor $2^l$, these positions can be mapped to the corresponding locations in the original image:
\begin{equation}
\label{eq:ositioins}
T_l = \{(i \times 2^l, j \times 2^l) \; | \; V_l(i,j) > \gamma\};
\end{equation}
these identified positions approximate the small objects’ locations, which are prone to being overlooked or inaccurately estimated.}

To address the non-uniform distribution of small objects in UAV images, the ASOE module utilizes the K-means {algorithm} to cluster $T_l$ into $\mathcal{N}$ distinct clusters. 
{The K-means {algorithm} achieves effective performance gains through K-value optimization, requiring only minimal parameter adaptation for cross-dataset deployment. 
This balances deterministic region control with adaptability to diverse data distributions, avoiding the parameter sensitivity of DBSCAN and the computational overhead of the mean-shift in real-time UAV applications. The cluster process can be represented as
\begin{equation}
    \{C_1,\ldots,C_\mathcal{N}\}=\text{K-means}(T_l,\mathcal{N}),
\end{equation}
where $C_k=\left\{(i,j)\in T_l \; | \; k=\arg\min_m\|\mathbf{x}-\mu_m\|^2\right\}$.
By determining the top-left and bottom-right coordinates of each cluster, we crop $\mathcal{N}$ subregions from the original image, defined as
\begin{equation}
    \{S_i,\ldots,S_\mathcal{N}\}=\text{CropByTLBR}(C).
\end{equation}
{Subsequently,} %MDPI:  Please confirm if there should be added indent
 we upscale the extracted $\mathcal{N}$ subregions and input them into a fine-grained detector, thereby enhancing the detection performance of small objects.}

\subsection{Dynamic Class-Balanced Copy--Paste (DCC)}
\label{subsec:dcc}

To address the challenge of class imbalance in UAV images,
resampling techniques like class-balanced sampling and repeat factor sampling are commonly employed.
However, these image-level resampling solutions elevate training costs. 
In contrast, the copy--paste method offers object-level resampling by copying and pasting objects and has shown significantly advanced object detection with ordinary images. 
However, its application to UAV images has been less effective, which is attributed to their complexity.
Therefore, we propose dynamic class-balanced copy--paste (DCC), which differs from traditional copy--paste solutions in considering both the diversity and position rationality of objects.

{Firstly, to mitigate overfitting caused by repetitive resampling of specific instances, we introduce a diversity augmentation (DA) strategy, leveraging dynamic memory banks with data augmentation techniques to ensure diverse and balanced training batches.
Specifically, each tail class is assigned a memory bank with a capacity of 10 to store instance info:
\begin{equation}
\label{eq:queue}
Q=\{\{(x_{i},b_{i})\}_{i\operatorname{=}1}^{10},\cdotp\cdotp\cdotp,\{(x_{i},b_{i})\}_{i\operatorname{=}1}^{10}\},
\end{equation}
where $x_i$ denotes a tail-class instance image, and $b_i$ represents its bounding box coordinates. 
During training, when the ASOE identifies a subregion $S$ containing small objects, all tail-class instances within this region are enqueued via a first-in-first-out (FIFO) \mbox{replacement policy:}
\begin{equation}
    Q=\mathrm{FIFO}(Q,S),
\end{equation}
where each instance image is cropped with $1.5\times$ bounding box expansion to preserve contextual semantics. Object-level resampling involves strategically pasting tail-class instances into subregions. We first sample an instance $(x,b)$ from $Q$ and then apply data augmentation techniques:
\begin{equation}
    (x^\prime,b^\prime)=\mathcal{A}(x,b),
\end{equation}
where $\mathcal{A}(\cdot)$ denotes data augmentation techniques such as {shift--scale--rotate} %MDPI: Please confirm if the italics are necessary; if not, please remove them. The following highlights are the same.
and {random brightness contrast}, and $(x^\prime,b^\prime)$ is the augmented instance for pasting.

Secondly, to make the pasting position more reasonable, we utilize a dynamic search (DS) strategy combined with ASOE clustering cues. Specifically, in ASOE, besides obtaining a subregion $S$ for fine-grained detection, we can also obtain its clustering center $(X,Y)$, which represents the distribution center of objects in each subregion. The ideal paste position should be closer to the cluster center and maintain no overlap with existing objects, so DCC dynamically searches a suitable position for $(x^\prime,b^\prime)$ by performing the {BFS algorithm}~\citep{bundy1984breadth} from the cluster center:
\begin{equation}
    (X^P,Y^P)=\mathrm{BFS}(S,(X,Y),mask,b^{\prime}),
\end{equation}
\begin{equation}
    mask(i,j) = 
\begin{cases}
1, & \text{if GT object exists at position } (i,j) \\
0, & \text{otherwise}
\end{cases} \quad \forall i,j
\end{equation}
where the width and height of $b^\prime$ are employed as the adaptive search step size, and the binary $mask$ enables rapid validation of an object's presence in $S$.
Afterward, $x^\prime$ is pasted into the subregion $S$, and the label set is expanded with $b^\prime$. Through these operations, we can dynamically and appropriately enhance both the quantity and quality of tail-class instances, effectively balancing their contribution during training.}

\subsection{Training and Inference Details}
\label{subsec:train-infer-details}
In the training process, AD-Det integrates ASOE and DCC to independently train the coarse and fine-grained detectors. 
To achieve coarse detection and extract the small object feature map, the coarse detector is first trained using all original images, with its training loss function defined as in Equation~\eqref{eq:coarse-loss}.
\begin{equation}
\label{eq:coarse-loss}
\mathcal{L}_{coarse}=\mathcal{L}_{cls}(x^g,y^g_{cls})+\mathcal{L}_{reg}(x^g,y^g_{reg}),
\end{equation}
where $\mathcal{L}_{cls}$ and $\mathcal{L}_{reg}$ denote classification loss and regression loss, and $\{x^g, y^g\}$ denote the original training set. 
Then, we conduct the coarse detector to traverse the original training set. 
For each sample, ASOE extracts small object subregions and combines with DCC to augment tail classes. 
This process yields the fine-grained training set $\{x^l, y^l\}$; the detailed {algorithm} is illustrated in {Algorithm}~\ref{alg:gfs}, and the training loss function for the fine-grained detector can be represented as Equation~\eqref{eq:fine-loss}.
\begin{equation}
\label{eq:fine-loss}
\mathcal{L}_{fine}=\mathcal{L}_{cls}(x^l,y^l_{cls})+\mathcal{L}_{reg}(x^l,y^l_{reg}),
\end{equation}
\vspace{-6pt}

\begin{algorithm}[H]
    \caption{{The} %MDPI: We moved the algorithm below the first mention, please confirm
 generation of fine-grained subregions}
    \label{alg:gfs}
%\SetAlgoLined
%\SetAlgoNoLine %去掉竖线
% \SetKwData{Left}{left}
% \SetKwData{This}{this}
% \SetKwData{Up}{up}
% \SetKwFunction{Union}{Union}
% \SetKwFunction{FindCompress}{FindCompress} 
% \SetKwInOut{Input}{input}
% \SetKwInOut{Output}{output}
\SetKwFunction{Filter}{Filter}
\SetKwFunction{Kmeans}{Kmeans}
\SetKwFunction{CropByTLBR}{CropByTLBR}
\SetKwFunction{DCC}{DCC}
	\KwIn{Normalized high-resolution heatmap $V$, threshold $\gamma$, the number of subregions $\mathcal{N}$, the state of $Training$ }
	\KwOut{Subregions $S$ }
	 \BlankLine
        $Q\gets \{\}$\;
        $T\gets$ \Filter{$V, \gamma$}\;
        \For{$(i,j)$ in $T$}{
            $(i, j)\gets (i \times 2^l, j \times 2^l)$\;
        }
        {
        $C_1...C_N \gets$ \Kmeans{$T, \mathcal{N}$}\;
        \For{$C_i$ in $C$}{
            $S_i \gets$ \CropByTLBR{$C_i$}\;
            \If{$Training$ is $True$}{
                $S_i \gets$ \DCC{$S_i, Q$}\;
            }
        }}
        $S \gets \bigcup S_i$\;
        \Return{$S$}\;
 	 
\end{algorithm}

In the inference process, due to the fine-grained detector’s ability to detect subregions with small objects and long-tail distributions, AD-Det exclusively relies on ASOE to acquire related subregions.
Let $G$ denote the input UAV image, and the final detection result $D_{final}$ can be formulated as in Equation~\eqref{eq:res-gen}.
\begin{equation}
\label{eq:res-gen}
D_{final}=NMS(\,Coarse(G),\ \bigcup_{i=1}^\mathcal{N}Fine(S_i)),
\end{equation}
where $Coarse(\cdot)$ and $Fine(\cdot)$ denote coarse and fine-grained detector, respectively.
$S_i$ represents the subregions extracted by ASOE, and NMS stands for non-maximum suppression~operation.

\section{Experiments}
\subsection{Experimental Setup}
\textbf{{Datasets.}} We evaluate our AD-Det on two publicly available datasets for UAV image object detection: VisDrone~\citep{zhu2021detection} and UAVDT~\citep{du2018unmanned}.
We summarize the details on these two datasets in Table~\ref{tab:datasets-detail}.

VisDrone is a challenging large-scale dataset captured by various camera devices using multiple UAVs. 
This dataset contains data collected from several Chinese cities, covering different weather conditions and scenarios.
Ten predefined categories, including {pedestrian, car, van, etc.}, are manually annotated with bounding boxes.
% VisDrone contains 6471 training images, 548 validation images, and 3190 testing images.
There are {6471} %MDPI: Commas are only used for numbers with five or more digits. We have removed them in four-digit numbers. Please confirm.
 images in the VisDrone for training, complemented by 548 images in the validation set, and a total of \mbox{3190 images} for testing.
The maximal resolution for each image is $2000\times1500$ pixels.
%% Lancer
%% 数据集划分，数据处理方式
Given that the testing set is not publicly available, we follow ClusDet~\citep{yang2019clustered} and DMNet~\citep{li2020density} to train the model on the training set while evaluating it on the validation set.

\begin{table}[H]
  % \centering
  \caption{Dateset description of VisDrone and UAVDT. 
  % "val/test" denotes the number of images on validation set of VisDrone and testing set on UAVDT
  }
    \begin{tabularx}{\textwidth}{CCCCC}
    \toprule
    % \multirow{2}[4]{*}{Dataset} & \multicolumn{2}{c}{\# of }  & \multirow{2}[4]{*}{resolution} & \multirow{2}[4]{*}{classes} \\  \cline{2-3}
    %      & train   & test  &       &  \\
    \textbf{Dataset}	& \textbf{Train}	& \textbf{Test} & \textbf{Resolution} & \textbf{Classes}\\
    \midrule
    VisDrone & 6471    &   548   &  $2000\times1500$  & 10 \\ 
    UAVDT &  23258        &   15069    &   $1080\times540$    & 3 \\
    \bottomrule
    \end{tabularx}%
  \label{tab:datasets-detail}%
\end{table}%

UAVDT is a large-scale vehicle detection and tracking dataset for UAV scenarios, which is released by UCAS.
This dataset contains data collected from complex environments with different weather conditions, occlusion, and flying heights.
% Three predefined categories (\textit{car, truck, and bus}) are manually annotated with bounding boxes. 
Bounding boxes have been manually annotated for three predefined categories, including {car, truck, and bus}.
It contains {23,258} %MDPI: We added commas to separate out the thousands for numbers with five or more digits. Please confirm.
 and \mbox{15,069 images} for training and testing, respectively, and the resolution is $1080\times540$ pixels for each image.
% We train the model on the training set and evaluate it on the testing set.

\textbf{{Evaluation metrics.}
} Following the evaluation protocols outlined in MS COCO~\citep{lin2014microsoft}, we employ average precision (AP) as our primary metric, spanning various categories and IoU thresholds. 
The metrics are succinctly outlined as follows: 
\begin{itemize}
    \item AP: average precision calculated across all categories, considering IoU values within the range of [0.5, 0.95] at intervals of 0.05. 
    \item AP\_50, AP\_75: average precision computed across all categories using individual IoU thresholds of 0.5 and 0.75, respectively.
    \item AP\_S, AP\_M, AP\_L: average precision calculated on object sizes small (less than $32^2$), medium (from $32^2$ to $96^2$), and large (greater than $96^2$).
\end{itemize}

\textbf{{Implementation details.}
} We implement AD-Det using the MMDetection toolbox~\citep{mmdetection} with PyTorch~\citep{paszke2019pytorch} as the basic framework. 
The model is trained on a system equipped with 2*Intel Silver 4210R CPU and 2*NVIDIA A4000 GPU. 
The baseline detection network employed is GFL~\citep{li2020generalized} with four uniform cropping parts.

%\noindent
Training phase:
For the training procedure on VisDrone and UAVDT, as~\citep{du2023adaptive} does, the input image size is configured as $1333\times800$ pixels for VisDrone and $1024\times540$ pixels for UAVDT, with both coarse and fine-grained detectors. 
Concerning hyperparameters in the ASOE, the maximum subregion number ($\mathcal{N}$) is set to 4 and 3 on VisDrone and UAVDT, respectively (as discussed in Section~\ref{exp:ablation}).
In the DCC module, all categories except \textit{pedestrian}, \textit{people}, and \textit{car} are considered tail classes on VisDrone, while \textit{truck} and \textit{bus} are considered tail classes on UAVDT.
% \textcolor{red}{In the DCC module, all categories except \textit{pedestrian}, \textit{people} and \textit{car} are considered as tail classes on VisDrone, while \textit{truck} and \textit{bus} are considered as tail classes on UAVDT.}
All models undergo 12 epochs of training using an SGD optimizer, with $momentum = 0.9$, $weight \, decay = 0.0001$,  
and $initial \, learning \, rate = 0.01$. 
AD-Det is trained with a linear warm-up strategy and undergoes decay by a factor of 10 at epochs 8 and 11.

%\noindent
In the testing phase, the input image size and hyperparameters in the ASOE remain consistent with the training phase unless otherwise specified. 
In the process of fusing detection, the non-max suppression (NMS) threshold and max detection number are set to 0.5 and 500, respectively, across all datasets.

\subsection{Comparison with Representative Solutions}
\textbf{{Results on VisDrone.}} We compare AD-Det with other SOTA solutions on VisDrone, and the results on AP, AP\_50, and AP\_75 are listed in Table \ref{tab:representative}. 
Our baseline method employs GFL with ResNet-50 / 101, and ResNeXt-101 as backbones.
From the table, we have the following observations.

\textbf{{(1) Our proposed method achieves consistent improvement upon all the compared solutions.}}%MDPI: Please confirm if keep noindent format.The following highlights are the same 
Our model improves the detection performance to 37.5\%, 60.9\%, and 39.2\% in terms of AP, AP\_50, and AP\_75, surpassing all compared methods.
{Among the compared solutions, ClusDet, DMNet, CDMNet, GLSAN, CZDet, AMRNet, and YOLC employ a coarse-to-fine strategy, which is similar to our approach. 
Even compared to CZDet, the best solution within them, our model also increases performance by 3.1\%, 1.2\%, and 4.6\% in AP, AP\_50, and AP\_75, respectively.}

\textbf{{(2) Despite using weaker backbones, our method surpasses or rivals the performance of all compared methods.}}
As illustrated at the bottom of Table~\ref{tab:representative}, although we employed a relatively weak and lightweight backbone ResNet-50, our model continues to be highly competitive. 
Our method exhibits substantial improvements in AP metrics compared to other methods, even with stronger backbones.
{{For example, compared to YOLC, which used ResNeXt-101 as the backbone, our method, with ResNet-50 as the backbone, nonetheless managed to increase performance by 2.2\%, 1.4\%, and 3.4\% in AP, AP\_50, and AP\_75, respectively.}}

{To further evaluate the accuracy and computational complexity of AD-Det, we present a comparison with existing solutions based on AP, parameter counts (Params), floating-point operations (FLOPs), and inference speed (s/img), as shown in Table~\ref{tab:complexity}. 
Under the same backbone architecture of ResNeXt-101, AD-Det demonstrates superior improvements in AP, Params, FLOPs, and inference speed compared to ClusDet, DMNet, and GLSAN. 
Although our method exhibits marginally higher FLOPs and slightly slower inference speed than CZDet and YOLC, it achieves higher detection accuracy and a significant reduction in Params. These results highlight that our approach delivers competitive performance with a better accuracy--speed trade-off, fulfilling practical engineering requirements for real-world deployment.}

\begin{table}[H]
\small
\caption{{{Comparison} %MDPI: We moved the table below the first mention, please confirm
 of AD-Det with SOTA approaches using AP, AP\_50, and AP\_75 on the validation set of VisDrone. 
  The results of comparative experiments are drawn from the corresponding literature. ``*'' denotes flip augment inference and bold values indicate the best results.\label{tab2}}}
%\isPreprints{\centering}{% This command is only used for ``preprints''.
	\begin{adjustwidth}{-\extralength}{0cm}
%\isPreprints{\begin{tabularx}{\textwidth}{CCCC}}{% This command is only used for ``preprints''.
		\begin{tabularx}{\fulllength}{p{3.2cm}CCCCCCCCC}
			\toprule
			% \textbf{Method}	& \textbf{Title 2}	& \textbf{Title 3}     & \textbf{Title 4}\\
            \multirow{2}[2]{*}{\textbf{Method}} & \multicolumn{3}{c}{\textbf{ResNet-50}} & \multicolumn{3}{c}{\textbf{ResNet-101}} & \multicolumn{3}{c}{\textbf{ResNeXt-101}} \\
\cmidrule{2-10}      & \textbf{AP} & \textbf{AP\_50} & \textbf{AP\_75} & \textbf{AP} & \textbf{AP\_50} & \textbf{AP\_75} & \textbf{AP} & \textbf{AP\_50} & \textbf{AP\_75} \\
            
			\midrule
Faster R-CNN ~\citep{ren2015faster} & 21.4 & 40.7 & 19.9 & 21.4 & 40.7 & 20.3 & 21.8 & 41.8 & 20.1 \\
    ClusDet~\citep{yang2019clustered} & 26.7 & 50.6 & 24.7 & 26.7 & 50.4 & 25.2 & 28.4 & 53.2 & 26.4 \\
    DMNet~\citep{li2020density} & 28.2 & 47.6 & 28.9 & 28.5 & 48.1 & 29.4 & 29.4 & 49.3 & 30.6 \\
    CDMNet~\citep{duan2021coarse} & 29.2 & 49.5 & 29.8 & 29.7 & 50.0 & 30.9 & 30.7 & 51.3 & 32.0 \\
    GLSAN~\citep{deng2021global} & 30.7 & 55.4 & 30.0 & 30.7 & 55.6 & 29.9 & - & - & - \\
    AMRNet~\citep{wei2020amrnet} & 31.7 & 52.7 & 33.1 & 31.7 & 52.6 & 33.0 & 32.1 & 53.0 & 33.2 \\
    CZDet~\citep{meethal2023cascaded} & 33.2 & 58.3 & 33.2 & 34.4 & 59.7 & 34.6 & - & - & - \\
    YOLC~\citep{liu2024yolc} & 31.8 & 55.0 & 31.7 & - & - & - & 33.7 & 57.4 & 33.8  \\
    \midrule
    AD-Det & 35.3 & 57.9 & 36.6 & 36.1 & 58.9 & 37.6 & 37.0 & 60.3& 38.3 \\
    AD-Det * & \textbf{{35.9} %MDPI: Please add an explanation for the use of bold in the table footer. If the bold is unnecessary, please remove it. The following highlights are the same.
} & \textbf{58.8} & \textbf{37.2} & \textbf{36.6} & \textbf{59.7} & \textbf{38.2} & \textbf{37.5} & \textbf{60.9} & \textbf{39.2} \\
			\bottomrule
		\end{tabularx}
%		\isPreprints{}{% This command is only used for ``preprints''.
	\end{adjustwidth}
	% \noindent{\footnotesize{* Tables may have a footer.}}
    \label{tab:representative}
\end{table}
\vspace{-12pt}

% \begin{table}[H] 
% %\tablesize{\small}
% \caption{This is a table caption. Tables should be placed in the main text near to the first time they are~cited.\label{tab1}}
% %\isPreprints{\centering}{} % Only used for preprints
% \begin{tabularx}{\textwidth}{CCC}
% \toprule
% \textbf{Title 1}	& \textbf{Title 2}	& \textbf{Title 3}\\
% \midrule
% Entry 1		& Data			& Data\\
% Entry 2		& Data			& Data \textsuperscript{1}\\
% \bottomrule
% \end{tabularx}
% \noindent{\footnotesize{\textsuperscript{1} Tables may have a footer.}}
% \end{table}

\begin{table}[H] 
    % \centering
    \caption{{Comparison of accuracy and complexity with existing solutions using AP, Params, FLOPs, and s/img on the validation set of VisDrone.}}
    \label{tab:complexity}
    
    \begin{tabularx}{\textwidth}{p{2.5cm}p{2.1cm}CCCC}
    \toprule
        
        \textbf{Method} & \textbf{Backbone}  &  \textbf{AP}   & \textbf{Params(M)} & \textbf{FLOPs (G)} &\textbf{s/img}  \\ 
        \midrule
        ClusDet~\citep{yang2019clustered} & ResNeXt-101 & 28.4 & 180.95 & 1647.25 &0.759 \\
        DMNet~\citep{li2020density} & ResNeXt-101 & 29.4  &  228.43 &  4492.18 & 0.957  \\ 
        GLSAN~\citep{deng2021global} & ResNet-101 & 30.7  &  590.86  &  2186.76 & 0.812 \\ 
        CZDet~\citep{meethal2023cascaded} & ResNet-101 & 34.4  & 120.32 & 1329.03 &0.593 \\ 
        YOLC~\citep{liu2024yolc} & ResNeXt-101 & 33.7  & 125.32 &  1245.25 &0.657 \\ \midrule
        % Ours & ResNet-50 & 35.9  & 64.10 & 1072.05  \\ \hline
        \multirow{3}[2] {*}{AD-Det}   & ResNet-50     & 35.3  & 64.10  & 1072.05 &0.514   \\
      & ResNet-101    & 36.1  & 102.10  & 1471.40 &0.615 \\
     & ResNeXt-101   & 37.0 & 101.36 & 1491.00 &0.701 \\
    \bottomrule
    \end{tabularx}
    
\end{table}

% \begin{table}[H] 
%     % \centering
%     \caption{Comparison of accuracy and complexity with existing solutions using AP, Params(M), FLOPs (G) and s/img on the validation set of VisDrone.}
%     \label{tab:complexity}
    
%     \begin{tabularx}{\textwidth}{p{2.5cm}p{2.1cm}CCC}
%     \toprule
        
%         \textbf{Method} & \textbf{Backbone}  &  \textbf{AP}   & \textbf{Params(M)} & \textbf{FLOPs (G)}   \\ 
%         \midrule
%         ClusDet~\citep{yang2019clustered} & ResNeXt-101 & 28.4 & 180.95 & 1647.25  \\
%         DMNet~\citep{li2020density} & ResNeXt-101 & 29.4  &  228.43 &  4492.18   \\ 
%         GLSAN~\citep{deng2021global} & ResNet-101 & 30.7  &  590.86  &  2186.76\\ 
%         CZDet~\citep{meethal2023cascaded} & ResNet-101 & 34.4  & 120.32 & 1329.03  \\ 
%         YOLC~\citep{liu2024yolc} & ResNeXt-101 & 36.3  & 125.32 &  1245.25  \\ \midrule
%         % Ours & ResNet-50 & 35.9  & 64.10 & 1072.05  \\ \hline
%         \multirow{3}[2]{*}{AD-Det}   & ResNet-50     & 35.9  & 64.10  & 1072.05    \\
%       & ResNet-101    & 36.6  & 102.10  & 1471.40  \\
%      & ResNeXt-101   & 37.5 & 101.36 & 1491.00  \\
%     \bottomrule
%     \end{tabularx}
    
% \end{table}

\textbf{{Results on UAVDT.}} Similar situations occur when our method is evaluated on UAVDT, 
and the detection accuracy compared with SOTA methods on the testing set of UAVDT is shown in Table \ref{tab:UAVDT}. 
To facilitate comparison, we utilize GFL with ResNet-50 backbone as our baseline.  
Compared with other SOTA methods, our method attains a new SOTA performance even with relatively weaker ResNet-50 as the backbone, achieving 20.1\% in AP, 34.2\% in AP\_50, and 21.9\% in AP\_75. 
{In comparison with the GLSAN, our model demonstrates increases of 3.1\%, 6.1\%, and 3.1\% in AP, AP\_50, and AP\_75, respectively.}

\begin{table}[H]
  % \centering
  \caption{{Comparison of AD-Det with SOTA approaches using AP, AP\_50, and AP\_75 on the testing set of UAVDT. 
  The results of comparative experiments are drawn from the corresponding literature. The best results are highlighted in bold.
  % {``-'' denotes not reported.} %MDPI:  There is no “-” in this table, please check and modify
}}
  % \textcolor{red}{Results of the comparative methods are sourced from relevant literature.} "-": not reported.}
    \begin{tabularx}{\textwidth}{p{3.3cm}p{2.6cm}CCC}
    \toprule
    \textbf{Method} & \textbf{Backbone} & \textbf{AP} & \textbf{AP\_50} & \textbf{AP\_75} \\
    \midrule
    
    Faster R-CNN~\citep{ren2015faster}  & ResNet-50 & 11.0 & 23.4  & 8.4  \\
    ClusDet~\citep{yang2019clustered}  & ResNet-50 & 13.7 & 26.5  & 12.5  \\
    DMNet~\citep{li2020density}  & ResNet-50 & 14.7  & 24.6  & 16.3  \\
    CDMNet~\citep{duan2021coarse} & ResNet-50 & 16.8 &29.1 & 18.5\\

    GLSAN~\citep{deng2021global} & ResNet-50 & 17.0  & 28.1  & 18.8  \\
    AMRNet~\citep{wei2020amrnet}  & ResNet-50 & 18.2  & 30.4  & 19.8  \\
    CZDet~\citep{meethal2023cascaded} & ResNet-50 & 18.9 & 30.2 & 20.3 \\
    YOLC~\citep{liu2024yolc}  & HRNet & 19.3 & 30.9  & 20.1 \\
    \midrule
    AD-Det  & ResNet-50  & \textbf{{20.1} %MDPI: Please add an explanation for the use of bold in the table footer. If the bold is unnecessary, please remove it. The following highlights are the same.
}  & \textbf{34.2}  & \textbf{21.9}  \\
    \bottomrule
    \end{tabularx}%
  \label{tab:UAVDT}%
\end{table}%

% % Table generated by Excel2LaTeX from sheet 'Sheet1'
% \begin{table}[H]
%   % \centering
%   \caption{Comparison of AD-Det with SOTA approaches using AP, AP\_50, and AP\_75 on the testing set of UAVDT. 
%   The results of comparative experiments are drawn from the corresponding literature. "-" denotes not reported.}
%   % \textcolor{red}{Results of the comparative methods are sourced from relevant literature.} "-": not reported.}
%     \begin{tabularx}{\textwidth}{p{3.3cm}p{2.6cm}CCC}
%     \toprule
%     \textbf{Method} & \textbf{Backbone} & \textbf{AP} & \textbf{AP\_50} & \textbf{AP\_75} \\
%     \midrule
    
%     Faster R-CNN~\citep{ren2015faster}  & ResNet-50 & 11.0 & 23.4  & 8.4  \\
%     ClusDet~\citep{yang2019clustered}  & ResNet-50 & 13.7 & 26.5  & 12.5  \\
%     DREN~\citep{zhang2019fully}      & ResNet-101 & 17.7 & - & - \\
%     DMNet~\citep{li2020density}  & ResNet-50 & 14.7  & 24.6  & 16.3  \\
%     GLSAN~\citep{deng2021global} & ResNet-50 & 17.0  & 28.1  & 18.8  \\
%     AMRNet~\citep{wei2020amrnet}  & ResNet-50 & 18.2  & 30.4  & 19.8  \\
%     AdaZoom~\citep{xu2023adazoom}  & ResNet-50 & 19.6 & 33.6 & 21.3 \\
%     PRDet~\citep{leng2023pareto}  & ResNet-50 & 19.8 & 34.1  & 21.3 \\
%     YOLC~\citep{liu2024yolc}  & HRNet & 19.3 & 30.9  & 20.1 \\
%     \midrule
%     AD-Det  & ResNet-50  & \textbf{20.1}  & \textbf{34.2}  & \textbf{21.9}  \\
%     \bottomrule
%     \end{tabularx}%
%   \label{tab:UAVDT}%
% \end{table}%

\subsection{Qualitative Analysis}
\begin{figure}[H]
%    \centering
    \includegraphics[width=\textwidth]{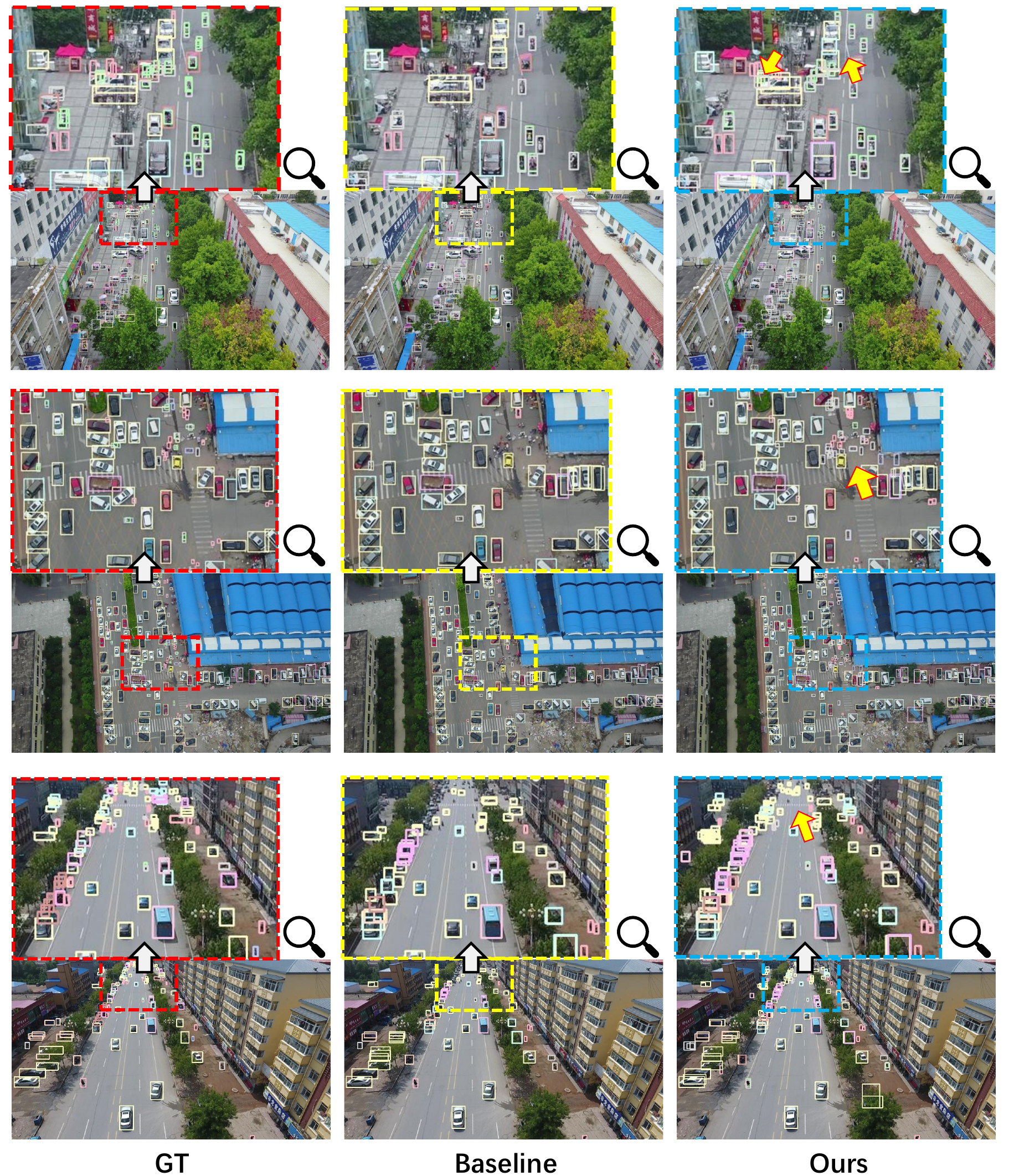}
    \caption{{Visualization} %MDPI: 1. Please confirm whether an explanation of the colors/arrows/box need to be added to the figure caption. 2. Please remove the non-English term from the figure or add a definition for it. 3.The contents of this figure are not legible. Please replace the image with one of a sufficiently high resolution (min. 1000 pixels width/height, or a resolution of 300 dpi or higher).
 of qualitative comparison between the baseline detector and our proposed method on VisDrone. The areas marked by the dashed boxes indicate that our approach excels in detecting small objects and tail categories.}
    \label{fig:quality_compare}
\end{figure}

To qualitatively contrast the performance, Figure~\ref{fig:quality_compare} depicts the detection results for both the baseline detector and our AD-Det.
% The qualitative comparison of detection visualization results between the baseline detector and our proposed method is illustrated in Figure~\ref{fig:quality_compare}. 
{It is evident that our method surpasses the baseline detector and reaches a satisfactory detection accuracy.}
Specifically, as depicted in the dashed area of 
% in the areas marked by the dashed boxes of 
Figure~\ref{fig:quality_compare} and the corresponding zoom-in views, our approach excels in detecting small objects and tail categories. 
This superiority can be attributed to the contributions of ASOE and DCC, which collectively enhance UAV image object detection from two distinct perspectives.

\subsection{Ablation Study}
\label{exp:ablation}
We conduct extensive studies on VisDrone to validate the effectiveness of our key design, including
(1) the effectiveness of ASOE and DCC modules,
(2) the choice of key hyperparameters,
% (3) The choice of the focused feature layer,
% (4) The design of the DCC module.
(3) the selection of the base detector,
(4) the design of the ASOE module, and
(5) the design of the DCC module.
% i.e., Adaptive Small Object Enhancement (ASOE) and  Dynamic Class-balanced Copy-paste (DCC). 
For all ablation studies, we employ ResNet-50 as our backbone. 
% TODO 增加均匀裁切结果，还有参数变化

\begin{figure}[H]
    \centering
    \includegraphics[width=\textwidth]{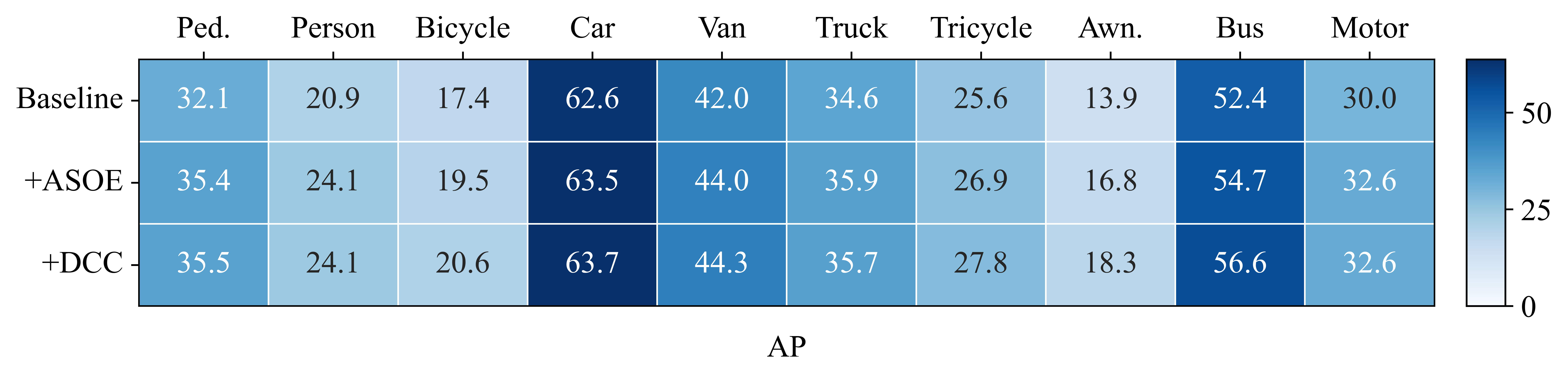}
    \caption{Ablation for ten fine-grained categories with AP. {Ped.}
 and {Awn.}
 denote {Pedestrian} and {Awning-tricycle}. ASOE and DCC indicate our proposed adaptive small object enhancement and dynamic class-balanced copy-paste method.}
    \label{fig:class-ablation}
\end{figure}

\textbf{{The effectiveness of ASOE and DCC modules:}} 
As the key components of our method, ASOE and DCC are systematically incorporated into the model to evaluate their respective efficacies. 
Simultaneously, we demonstrate the improvements resulting from the integration of ASOE and DCC, emphasizing their mutual complementarity.
A comprehensive presentation of the ablation study is provided in Table \ref{tab:ablation} and Figure \ref{fig:class-ablation}, which illustrate the accuracy improvements across different metrics after integrating our key components.

\begin{table}[H]
  % \centering
  \caption{The impacts of the proposed modules on detection performance in VisDrone. 
  ASOE and DCC indicate our proposed adaptive small object enhancement and dynamic class-balanced copy--paste method. The best results are highlighted in bold.}

    %\isPreprints{\centering}{% This command is only used for ``preprints''.
	% \begin{adjustwidth}{-\extralength}{0cm}
%\isPreprints{\begin{tabularx}{\textwidth}{CCCC}}{% This command is only used for ``preprints''.
		\begin{tabularx}{\textwidth}{lCCCCCCC}
    \toprule
    
    \textbf{Method}          & \textbf{AP}   & \textbf{AP\_50}  & \textbf{AP\_75} & \textbf{AP\_S}   & \textbf{AP\_M}  & \textbf{AP\_L} & \textbf{s/img} \\
    
    \midrule
          % Baseline      & 29.3  & 48.2  & 30.2  & 19.4 & 42.1 & 45.2 &0.131  \\
          Baseline & 33.1 &55.0 &34.0 &24.3 &44.1 &44.9 & 0.706\\
        Baseline + ASOE      & 35.3  & 58.0  & 36.5 & 27.5 & 44.8 & \textbf{45.4} & 0.758 \\
      Baseline + ASOE + DCC          & \textbf{{35.9} %MDPI: Please add an explanation for the use of bold in the table footer. If the bold is unnecessary, please remove it. The following highlights are the same.
}  & \textbf{58.8} & \textbf{37.2} & \textbf{28.0} & \textbf{45.7} & 45.2 &0.758 \\
    \bottomrule
		\end{tabularx}
%		\isPreprints{}{% This command is only used for ``preprints''.
	% \end{adjustwidth}

  \label{tab:ablation}%
\end{table}%

As illustrated in Table \ref{tab:ablation}, the incorporation of ASOE results in a notable improvement of 2.2\% in AP and 3.0\% in AP\_S over the baseline method. 
% This improvement can be attributed to the precise localization of small object regions by ASOE, thereby enhancing the performance in the fine-grained detection phase, especially for small-sized objects. 
Similarly, DCC contributes to an enhancement in detection performance, elevating it from 35.3\% to 35.9\% in AP. 
% {The traditional uniform cropping (UC) method effectively improve AP, substantiating the importance of focusing on subregions.
% However, UC is limited to enlarging the cropped regions to a fixed scale, which does not meet the adaptive detection requirements in large scenarios. 
% In contrast, ASOE can adaptively focus based on object distribution. 
Compared to baseline with identical parameter settings, our method achieves a notable increase in accuracy, but only increases time consumption slightly.

As illustrated in Figure~\ref{fig:class-ablation}, a noteworthy observation is the consistent enhancement in results for objects in the tail categories, especially for \textit{bicycles, tricycles, awn.}, and \textit{buses}, which improved by 0.9\%$\sim$1.9\%, underscoring the ability of DCC to augment detection performance reliably. 
Significantly, the addition of complementary samples through DCC does not adversely affect the performance of other categories.

\begin{figure}[H]
%    \centering
    \includegraphics[width=\textwidth]{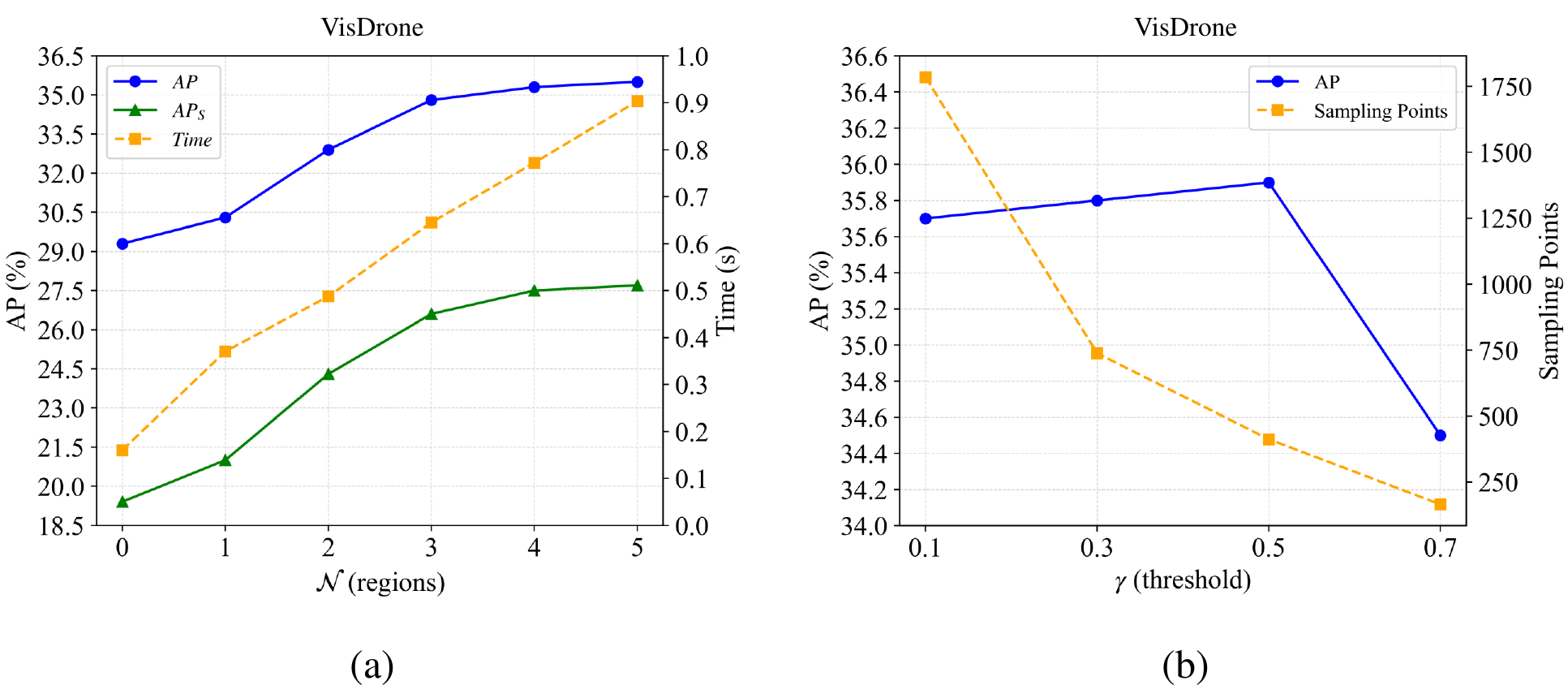}
    \caption{(\textbf{a}) AP, AP\_S and inference time curves on different numbers of subregions $\mathcal{N}$ on the validation set of VisDrone. (\textbf{b}) AP and sample point curves on different confidence thresholds' $\gamma$ values in Equation~\eqref{eq:ositioins} on the validation set of VisDrone. The sample points denote the points utilized for clustering in ASOE.}
    \label{fig:N and Threshold}
\end{figure}

\textbf{{The choice of key hyperparameters:}} We analyze the impact of two key hyperparameters herein---the number of subregions $\mathcal{N}$ in Figure~\ref{fig:N and Threshold}a and the confidence threshold $\gamma$ in Figure~\ref{fig:N and Threshold}b.

The number of subregions $\mathcal{N}$ is a pivotal parameter that significantly influences the detection speed and accuracy. 
The effects of different values of $\mathcal{N}$ on detection time and accuracy are summarized in Figure~\ref{fig:N and Threshold}a. 
This subfigure indicates that as the number of subregions increases, the network's detection accuracy improves, but with an associated increase in detection time. 
Considering the trade-off between accuracy and efficiency, this study selects $\mathcal{N}$ = 4 as the optimal number of subregions. 
We conduct a similar experiment on UAVDT and find that $\mathcal{N}$ = 3 is the optimal choice for UAVDT.

The confidence threshold $\gamma$ determines the quality of the subregions (Equation~\eqref{eq:ositioins}). 
We experiment with different $\gamma$ values in \{0.1, 0.3, 0.5, 0.7\}. 
In particular, we quantify sample points and their corresponding detection accuracy. 
As depicted in Figure~\ref{fig:N and Threshold}b, with an increase in $\gamma$, the number of sample points decreases, facilitating a proportional acceleration in clustering speed. 
However, the accuracy benefits are not linear with higher $\gamma$ values. 
When $\gamma$ = 0.7, an excessive number of points are disregarded, leading to a significant decline in detection accuracy. 
Therefore, we select $\gamma$ = 0.5 as the \mbox{threshold parameter.}

\textbf{{The selection of a base detector:}}
In Table~\ref{tab:base-detector}, we present a comparison of GFL with several leading base detectors, where GFL achieves superior performance. As indicated by AP\_S, both GFL and CenterNet are effective in small object detection, but GFL exhibits more balanced accuracy across various object scales. In contrast to FCOS and YOLOv8, GFL outperforms both in AP and AP\_S. Therefore, we choose GFL as the base detector of AD-Det.

\begin{table}[H]
    % \centering
    \caption{Comparison of GFL with leading base detectors. The best results are highlighted in bold.}
    
    \begin{tabularx}{\textwidth}{lCCCCCC}
    \toprule
        \textbf{Method}          & \textbf{AP}   & \textbf{AP\_50}  & \textbf{AP\_75} & \textbf{AP\_S}   & \textbf{AP\_M}  & \textbf{AP\_L}  \\
        \midrule
        FCOS~\citep{tian2019fcos} & 26.7 & 45.2 & 27.1 & 17.4 & 38.8 & 45.1  \\ 
        CenterNet~\citep{zhou2019objects} & 27.2 & 49.8 & 25.8 & 19.4 & 38.7 & 39.2  \\ 
        YoloV8~\citep{wang2023yolov7} & 26.1 & 43.9 & 26.4 & 15.8 & 39.2 & \textbf{52.1}  \\ 
        GFL~\citep{li2020generalized} & \textbf{{29.3} %MDPI:Please add an explanation for the use of bold in the table footer. If the bold is unnecessary, please remove it. The following highlights are the same. 
} & \textbf{48.2}& \textbf{30.2} & \textbf{19.4} & \textbf{42.1} & 45.2  \\ \bottomrule
    \end{tabularx}
    \label{tab:base-detector}
\end{table}

\begin{figure}[H]
    \centering
    \includegraphics[width=\textwidth]{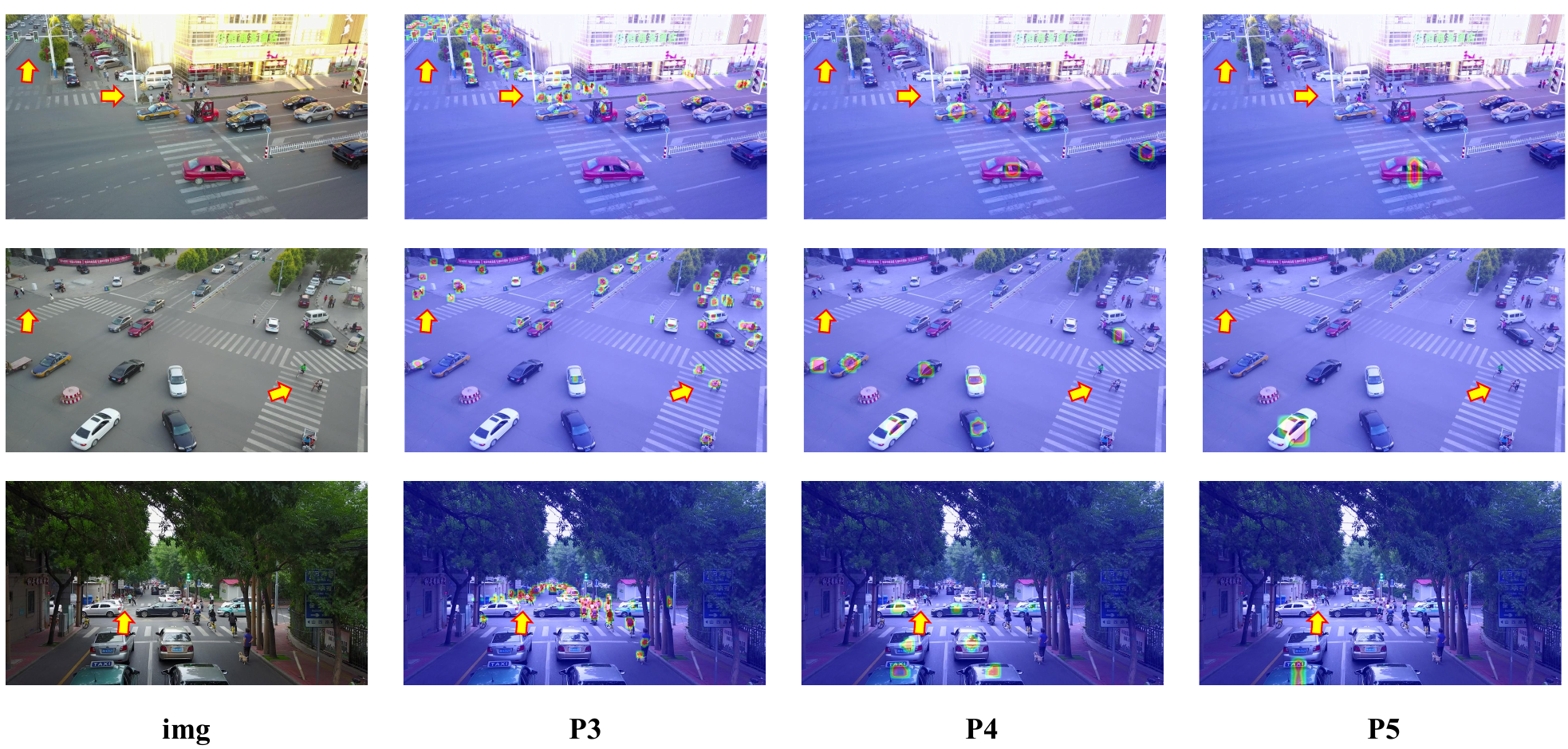}
    \caption{{Visualization} %MDPI: 1.Please confirm whether an explanation of the arrow/colors needs to be added to the figure caption. 2. Please remove the non-English term from the figure or add a definition for it.
 of interesting regions selected by ASOE using different feature maps. Arrows point to small objects.}
    \label{fig:heatmap}
\end{figure}

\textbf{{The design of the ASOE module:}}
In our ASOE module, we empirically select the lowest layer of the feature maps, namely $P_3$, as our primary focus~\citep{lin2017feature}. 
In Figure~\ref{fig:heatmap}, we visualize the regions of interest selected by ASOE using different feature maps. 
The visualization shows that the $P_3$ layer focuses more on small object regions, while higher layers such as $P_4$ and $P_5$ gradually focus on larger objects. 
To substantiate its efficacy, we conduct a straightforward comparison with the higher adjacent layer $P_4$ as an alternative focus. 
We summarize the results in Table \ref{tab:layer}. 
Experimental results indicate that the $P_3$ layer yields the highest performance, while clustering from higher-level layers such as $P_4$ would lead to performance degradation.
Note that AP and AP\_S undergo a sharp decrease from $P_3$ to $P_4$, highlighting our approach's effectiveness in detecting small objects within \mbox{higher-resolution layers.
}

\textbf{{The design of the DCC module:}}
DCC comprises two major components---diversity augmentation (DA) and dynamic search (DS). 
Based on ASOE, we verify these two components step by step. 
As shown in Table~\ref{tab:DCC-ablation}, when DA is applied to tail-class instances, the AP\_S and AP\_M are increased by 0.4\% and 0.7\%, respectively. 
% This is due to the fact that DA can enhance the diversity of instances in tail-class, which are dominated by small and medium-sized objects. 
When further combined with DS, the AP is increased to 35.9\%, which proves its effectiveness.
% the appropriate paste position can alleviate the inconsistency between instances and the background, thus conducting reasonable resampling and improving detection performance for tail-class.

\begin{table}[H]
  % \centering
  \caption{Ablation for our ASOE module. $P_3$ denotes the lowest layer of the feature maps, and $P_4$ is the higher adjacent layer. The best results are highlighted in bold.}
    \begin{tabularx}{\textwidth}{lCCCCCC}
    \toprule
    \textbf{Method}          & \textbf{AP}   & \textbf{AP\_50}  & \textbf{AP\_75} & \textbf{AP\_S}   & \textbf{AP\_M}  & \textbf{AP\_L}  \\
    \midrule
    % Baseline   & 29.3       & 48.2       & 30.2 & 19.4 & 42.1 & 45.2  \\
    Baseline  & 33.1  &55.0 & 34.0 & 24.3 & 44.1 & 44.9 \\
    with $P_4$    & 30.4       & 50.9       & 30.9 & 21.3 & 42.2 & 45.4  \\
    with $P_3$ (ours) & \textbf{{35.3}  %MDPI:Please add an explanation for the use of bold in the table footer. If the bold is unnecessary, please remove it. The following highlights are the same. 
}       & \textbf{58.0}       & \textbf{36.5} & \textbf{27.5} & \textbf{44.8} & \textbf{45.4} \\
    \bottomrule
    \end{tabularx}%
  \label{tab:layer}%
\end{table}%
\vspace{-12pt}

\unskip
% \subsubsection{The choice of focused feature layer}

% Further, the traditional uniform cropping (UC) method effectively improve AP, substantiating the importance of focusing on subregions.
% However, UC is limited to enlarging the cropped regions to a fixed scale, which does not meet the adaptive detection requirements in large scenarios. 
% In contrast, ASOE can adaptively focus based on object distribution. 
% Compared to UC under identical hyperparameter settings, our method achieves a 2.2\% improvement in AP and a notable 3.2\% one in AP\_S.

\begin{table}[H]
  % \centering
  \caption{Ablation for our DCC module. DA and DS indicate diversity augmentation for tail classes and dynamic position search for tail classes, respectively. The best results are highlighted in bold.}
    \begin{tabularx}{\textwidth}{lCCCCCC}
    \toprule
    
         \textbf{Method}          & \textbf{AP}   & \textbf{AP\_50}  & \textbf{AP\_75} & \textbf{AP\_S}   & \textbf{AP\_M}  & \textbf{AP\_L}  \\
    \midrule
     ASOE     &      35.3  & 58.0  & 36.5 & 27.5 & 44.8 & \textbf{45.4}  \\
     ASOE + DA        & 35.6  & 58.5  & 36.9 & 27.9 & 45.5 & 45.3  \\
      ASOE + DA + DS    & \textbf{{35.9}  %MDPI:Please add an explanation for the use of bold in the table footer. If the bold is unnecessary, please remove it. The following highlights are the same. 
}  & \textbf{58.8} & \textbf{37.2} & \textbf{28.0} & \textbf{45.7} & 45.2  \\
    \bottomrule
    \end{tabularx}%
  \label{tab:DCC-ablation}%
\end{table}%

% \subsubsection{ The design of the DCC module}

% \section{Discussion}

\section{Conclusions}
In this paper, we propose AD-Det, a novel object detection framework for UAV images, which addresses the challenges of scale variations and class imbalance.
AD-Det consists of two key components: adaptive small object enhancement (ASOE) and dynamic class-balanced copy--paste (DCC). 
ASOE employs a high-resolution feature map to cluster regions containing small objects, followed by processing them for fine-grained detection. 
DCC performs reasonable object resampling by dynamically pasting tail classes around clustering centers obtained by ASOE. 
We conduct extensive experiments on VisDrone and UAVDT and demonstrate that our approach significantly outperforms existing competitive~alternatives. 

Although our proposed approach achieves satisfactory results in UAV image object detection, several limitations remain to be addressed in our future endeavors.
% there are still some limitations, which would need to be resolved in future work.
(1) ASOE achieves precise detection of small objects by leveraging global location cues, but the usage of global features is still limited. 
ASOE ignores the feature interaction between the global image and the local regions, which may help to better understand the scene's semantic knowledge. 
{(2) DCC conducts uniform sampling copy--paste, but there is room for improvement by considering instance-specific metrics and the complex relationships between classes for more effective hard example mining.}
% DCC performs dynamic class balancing on existing images, but ignores the more flexible way of constructing new images with tail-class instances.
In the future, we plan to explore more effective ways of handling UAV images under complex scenarios, including occlusion and background clutter issues.

%%%%%%%%%%%%%%%%%%%%%%%%%%%%%%%%%%%%%%%%%%
% \authorcontributions{For research articles with several authors, a short paragraph specifying their individual contributions must be provided. The following statements should be used ``Conceptualization, X.X. and Y.Y.; methodology, X.X.; software, X.X.; validation, X.X., Y.Y. and Z.Z.; formal analysis, X.X.; investigation, X.X.; resources, X.X.; data curation, X.X.; writing---original draft preparation, X.X.; writing---review and editing, X.X.; visualization, X.X.; supervision, X.X.; project administration, X.X.; funding acquisition, Y.Y. All authors have read and agreed to the published version of the manuscript.'', please turn to the  \href{http://img.mdpi.org/data/contributor-role-instruction.pdf}{CRediT taxonomy} for the term explanation. Authorship must be limited to those who have contributed substantially to the work~reported.}

\vspace{6pt} 

\authorcontributions{Conceptualization, Z.L. and S.L.; methodology, Z.L.; software, D.P. and Y.W.; validation, Z.L., S.L. and D.P.; formal analysis, Z.L. and S.L.; investigation, Z.L. and Y.W.; resources, Z.L.; data curation, Z.L.; writing---original draft preparation, Z.L. and S.L.; writing---review and editing, S.L. and W.L.; visualization, Z.L.; supervision, S.L. All authors have read and agreed to the published version of the manuscript.}

% \section*{Aknowledgments}
%\clearpage 
\funding{This research was funded by the Natural Science Foundation of Fujian Province of China youth project (No. 2023J05117); the Education and Scientific Research Project for Middle-aged and Young Teachers in Fujian Province (No. JAT220017); the National Natural Science Foundation of China (No. 62461026); and the Natural Science Foundation of Jiangxi Province (\mbox{No. 20232BAB203057}).}

\dataavailability{The datasets used in this study are all open-source: VisDrone is available at \url{https://github.com/VisDrone/VisDrone-Dataset} (accessed on 17 January 2025), and UAVDT can be accessed at \url{https://sites.google.com/view/grli-uavdt} (accessed on 17 January 2025).}

% \dataavailability{We encourage all authors of articles published in MDPI journals to share their research data. In this section, please provide details regarding where data supporting reported results can be found, including links to publicly archived datasets analyzed or generated during the study. Where no new data were created, or where data is unavailable due to privacy or ethical restrictions, a statement is still required. Suggested Data Availability Statements are available in section ``MDPI Research Data Policies'' at \url{https://www.mdpi.com/ethics}.} 

\conflictsofinterest{The authors declare no conflicts of interest.}

%%%%%%%%%%%%%%%%%%%%%%%%%%%%%%%%%%%%%%%%%%
%\isPreprints{}{% This command is only used for ``preprints''.
\begin{adjustwidth}{-\extralength}{0cm}

\reftitle{References}
\PublishersNote{}
%\isPreprints{}{% This command is only used for ``preprints''.
\end{adjustwidth}
\end{document}